\def\year{2022}\relax
\documentclass[letterpaper]{article} 
\usepackage{aaai22}  
\usepackage{times}  
\usepackage{helvet}  
\usepackage{courier}  
\usepackage[hyphens]{url}  
\usepackage{graphicx} 
\usepackage{natbib}  
\usepackage{caption} 
\DeclareCaptionStyle{ruled}{labelfont=normalfont,labelsep=colon,strut=off} 
\usepackage{algorithm}
\usepackage{algorithmic}

\usepackage{newfloat}
\usepackage{listings}
\floatstyle{ruled}
\newfloat{listing}{tb}{lst}{}
\floatname{listing}{Listing}
\usepackage[hyphens]{url}
\usepackage{hyperref}
\usepackage[hyphenbreaks]{breakurl}

\usepackage{booktabs}

\usepackage{amsmath}
\usepackage{mathtools}
\usepackage{cleveref}
\usepackage{multirow}
\usepackage{csquotes}

\DeclareMathOperator*{\argmax}{arg\,max}

\newcommand{\Substack}[1]{\hbox to 0pt{\hss $\substack{ #1}$\hss}}
\title{Adapting Document-Grounded Dialog Systems to Spoken Conversations using~Data Augmentation and a Noisy Channel Model}
\author{
    David Thulke,\equalcontrib\textsuperscript{\rm 1,2}
    Nico Daheim,\equalcontrib\textsuperscript{\rm 1}
    Christian Dugast,\textsuperscript{\rm 2}
    Hermann Ney\textsuperscript{\rm 1,2}\\
}
\affiliations{
    \textsuperscript{\rm 1} Human Language Technology and Pattern Recognition Group, RWTH Aachen University, Germany \\
    \textsuperscript{\rm 2} AppTek GmbH, Aachen, Germany \\
    \{thulke, daheim, ney\}@i6.informatik.rwth-aachen.de \\
    cdugast@apptek.com \\
}

\begin{document}

\maketitle

\begin{abstract}
This paper summarizes our submission to Task 2 of the second track of the 10th Dialog System Technology Challenge (DSTC10) \enquote{Knowledge-grounded Task-oriented Dialogue Modeling on Spoken Conversations}.
Similar to the previous year's iteration, the task consists of three subtasks: detecting whether a turn is knowledge seeking, selecting the relevant knowledge document and finally generating a grounded response.
This year, the focus lies on adapting the system to noisy ASR transcripts.
We explore different approaches to make the models more robust to this type of input and to adapt the generated responses to the style of spoken conversations.
For the latter, we get the best results with a noisy channel model that additionally reduces the number of short and generic responses.
Our best system achieved the 1st rank in the automatic and the 3rd rank in the human evaluation of the challenge.
\end{abstract}

\section{Motivation}
While research on document-grounded dialog systems has been focusing on written dialogs, many of their applications, such as voice assistants, are within the domain of spoken conversations.
However, the two domains are fundamentally different.
First of all, spoken conversations are more spontaneous and may include interruptions, repetitions, corrections, and other disfluencies. 
If the speech was transcribed automatically, errors propagated from automatic speech recognition (ASR) systems introduce further challenges.
Hence, it is not clear a priori whether a dialog system trained on written data would also fit spoken data sufficiently well.
Indeed, \citet{Gopalakrishnan2020AreNO} and \citet{kimHowRobustEvaluating2021} show that the performance of existing models trained on written conversations strongly degrades when evaluated on spoken data.

The second track of the 10th Dialog System Technology Challenge (DSTC10) hosts a shared task on "Knowledge-grounded Task-oriented Dialogue Modeling on Spoken Conversations".
In continuation of the first track of the DSTC9 challenge \cite{kimDomainAPIsTaskoriented2021}, three subtasks are proposed.
First, the dialog system has to identify user turns that seek knowledge not defined by an API and only available in the form of unstructured documents.
This task is called Knowledge-seeking Turn \textit{Detection}.
Next, the system has to retrieve a document and generate a response based on this document and the dialog context.
These tasks are called Knowledge \textit{Selection} and Response \textit{Generation}, respectively.
To address the aforementioned issues, this time evaluation is only done on automatic transcripts of spoken conversations including ASR errors to test the robustness of dialog models.
Only a small validation set of spoken conversations is provided to develop models.

We explore different methods to adapt document-grounded dialog systems to spoken conversations.
For the detection and selection subtasks, we explore simple text preprocessing techniques applied to the training data in order to make the models more robust to ASR transcripts.
In addition, we make use of the provided n-best lists to capture the uncertainties of the transcriptions.
For the generation subtask, we additionally focus on adapting the style of generated responses so that they are more naturally connected to the dialog.
Therefore, we experiment with a few-shot transfer using the DSTC10 validation data and explore approaches to integrate an external (ungrounded) response generation model trained on spoken conversations.
One of these approaches is a Noisy Channel reformulation of the generation task.
In addition to the integration of the external model, this formulation allows penalizing short and generic responses that are often generated by the baseline model.
We hypothesize that this increases the faithfulness of generated responses.

\section{Task Description}

The DSTC10 challenge does not provide a dedicated training set.
Rather, teams were encouraged to use the training set of the DSTC9 challenge consisting of 72,518 dialogs in written style.
This set is an augmented version of MultiWOZ 2.1 \cite{ericMultiWOZConsolidatedMultiDomain2020} where turns were added that are grounded in external knowledge documents as opposed to the existing API-based turns.
Documents were collected from FAQ pages and consist of a question-answer pair.
Thus, they are relatively short.
They span the four different domains: hotel, restaurant, train, and taxi. The first two are divided further into entities.
In the following we refer to the set of documents as knowledge base.
While the validation set makes use of the same documents and locality as the training set, the test set introduced a new locality - San Francisco - and unseen documents, some of which stem from a new domain called attraction.
Around half of the 4,181 dialogs from the test set were obtained again by augmenting the MultiWOZ dataset.
The other half was collected from human-to-human conversations covering touristic information in San Francisco.
Around a tenth of these conversations are from the spoken domain and thus have a similar style as the DSTC10 data.
However, the DSTC9 test data does not include ASR transcripts but rather human transcripts without errors.
The validation data of DSTC10 are the same 263 dialogs as the spoken part of the DSTC9 test data but transcribed with an ASR system.
The test data consists of 1,988 additional dialogs collected with the same locality.
The ASR system that was used to transcribe the validation and test data is described in \citet{kimHowRobustEvaluating2021}.
The system achieved a word error rate of 24,09\% resulting in strongly perturbed outputs.
The knowledge base is the same as in the DSTC9 test set.
For more details on the datasets, we refer to \citet{kimHowRobustEvaluating2021}.

Accordingly, the tasks of this challenge are also the same as the ones of its predecessor.
The three tasks are defined as follows:
\paragraph{Detection} In Knowledge-seeking Turn Detection, the system has to decide whether the current user turn is covered by the given API or whether it requires unstructured knowledge access, i.e. one of the FAQ documents contains the required information.
\paragraph{Selection} In Knowledge Selection, the system has to find the FAQ document $K^\prime$ that answers the last knowledge-seeking user turn $u_{T}$.
\paragraph{Generation} Finally, response generation is the task of generating an appropriate agent response $u_{T+1}$ based on the selected knowledge document $K^\prime$ and dialog context $u_1^T$.

\section{Methods}

\subsection{Text Preprocessing}

In contrast to the written training data, the ASR transcripts in the DSTC10 validation and test data are lower-cased and do not contain punctuation.
This creates a mismatch between the training and evaluation data.
One approach to solve this issue is to augment the ASR transcripts with punctuation and casing information.
The downside of this approach is that external modules for these tasks are required that may introduce further errors in the pipeline.

An alternative approach is to remove this information from the written text so that it becomes more similar to the ASR transcripts.
In addition to that, we write out numbers (e.g. \emph{42} $\mapsto$ \emph{forty two}) and spell out abbreviations (e.g. \emph{mm} $\mapsto$ \emph{millimeters}).

\subsection{Detection}

As in the baseline model proposed by \citet{kimDomainAPIsTaskoriented2020}, we model knowledge detection as a binary classification task.
Similar to previous work \cite{miGeneralizedModelsDomain2021, jinCanBeFurther2021}, we fine-tune RoBERTa-large \cite{liuRoBERTaRobustlyOptimized2019}.
Therefore, we add a simple classifier consisting of two linear layers on top of the first hidden state (corresponding to the begin of sentence token).
To limit the length of the input, we only pass the last three utterances to the model and additionally truncate the input sequence if it exceeds 384 tokens.

To adapt the model to the knowledge documents from the new domains and localities, we generate additional knowledge-seeking dialog samples based on the documents in our knowledge base.
Therefore, for each document, we randomly select one dialog from the original MultiWOZ corpus in the same domain, replace the entity in the document with an entity from the dialog, and add the questions of the (faq) document as a new knowledge-seeking turn.
This way, we add 16,675 new samples to the training data.

To make use of the n-best list provided by the ASR system, we pass each ASR hypothesis to our model and experiment with two different strategies.
The \emph{best} strategy just selects the highest score of all hypotheses and the \emph{weighted} strategy calculates the weighted sum of all scores based on the (renormalized) probabilities of the ASR hypotheses.
Even though the \emph{weighted} strategy is the mathematically more sound option, as it treats the ASR hypotheses as a latent variable, we observe the highest F1 scores on the validation data with the \emph{best} strategy.

For our final model, we use an ensemble of different model variants and fine-tune the decision threshold on the DSTC10 validation data.

\subsection{Selection}

The goal of the selection subtask is to find the most relevant document from the knowledge base given a dialog.
The baseline system models this as a relevance classification task, i.e. for each pair the model predicts whether it is relevant or not and the document with the highest relevance score is chosen.
As this requires a full pass through the model for each pair of context and document, the method becomes increasingly inefficient for larger knowledge bases.
To avoid this, \citet{thulkeEfficientRetrievalAugmented2021} propose a Hierarchical Selection model that is identified as a good tradeoff between efficiency and performance.
To reduce the search space, they first use an entity model \(p_E\) to identify the most relevant domain and entity and then another document model \(p_D\) to identify the most relevant document of the selected entity.
In the following, we discuss the extensions we made to this approach.

First, since the entity selection task is similar to other task-oriented dialog tasks, it allows us to make use of additional training data.
Therefore, we collect those dialogs whose states contain entities of the relevant domains from Taskmaster-2\footnote{\url{https://github.com/google-research-datasets/Taskmaster/tree/master/TM-2-2020}} \cite{byrneTaskmaster1RealisticDiverse2019} and the DSTC10 Track 2 Task 1 validation data \cite{kimHowRobustEvaluating2021}.
In total 4,499 new training instances were added.

To train both models, we provide the reference entity/document as a positive sample and sample three negative samples in each epoch.
For the entity selection model, we randomly sample one entity with a different domain and two entities with the same domain as the negative samples.
For the document selection model, we sample three documents of the same entity as negative samples.

In the original variant proposed by \citeauthor{thulkeEfficientRetrievalAugmented2021}, a greedy search method was used.
Specifically, during the document selection, only documents of the most relevant entity are considered.
Instead, we propose to consider all entities whose entity relevance score \(p_E(r|e,u_1^T)\) is within a threshold \(t \leq 1\) of the most relevant entity \(\hat e\):
\begin{align}
    \hat k = \argmax_{\Substack{k = (e, d)\\p(e \mid  u_1^T) > t \cdot p(\hat e \mid u_1^T)}} \,\, p_E(e \mid u_1^T)^\gamma \cdot p_D(d \mid e, u_1^T)
\end{align}
Additionally, we add a scaling factor \(\gamma\) to control the influence of both models on the final selection.

To make use of the ASR n-best lists, we experimented with the same strategies as for the detection task.
Finally, for the DSTC10 validation and test data, we can further reduce the search space by only considering entities from the San Francisco locality.

\subsection{Generation}
To maintain a fluent conversation, generated responses should be naturally connected to the context of the dialog and thus match the style of preceding utterances.
Hence, we study different methods to encourage the model to generate answers in spoken style with either no or only a few in-domain samples.
In the following let $w \coloneqq u_{T+1}$ denote the generated response and $w_n$ its n-th token.
An intuitive method is to train the standard model
$$
p\left(w_n  \mid w_1^{n-1}, u_1^T, K^\prime \right) \text{, }
$$
to which we will refer as \emph{direct model} (\emph{dm}) in the following, on additional spoken dialogs.
While the model could infer the style of the dialog already from the context $u_1^T$, we further introduce a style token as a form of explicit conditioning.
Then, the model becomes
$$
p\left(w_n  \mid w_1^{n-1}, u_1^T, K^\prime, s \right) \text{, }
$$
where $s$ is a special $\langle \text{written} \rangle$ or $\langle \text{spoken} \rangle$ token that is added to the vocabulary.
Since only a few samples from the DSTC10 validation set are available to train this model, we seek methods that facilitate domain adaptation without requiring grounded dialogs, as such data is more readily available.
First, we try out a \emph{shallow fusion} \cite{gulcehre2015using}, i.e. a log-linear model combination, of the direct model and an \emph{ungrounded} response generation model (\emph{lm}) trained on spoken conversations.
We combine the models locally such that the response is generated according to
\begin{align*}
    p\left(w_n \mid w_1^{n-1}, u_1^T, K^\prime \right) &\propto p_{\text{dm}}\left(w_n \mid w_1^{n-1}, u_1^T, K^\prime\right) \\ &\cdot p_{\text{lm}}\left(w_n \mid w_1^{n-1}, u_1^T\right)^\lambda \text{,}
\end{align*}
where $\lambda$ is a scaling factor that is tuned on the validation data.
Further, to reduce the influence of the written style from the training data, we experiment with subtracting the output of a response generation model trained only on the written in-domain data without document grounding (\emph{ilm}).
Here, we draw inspiration from the Density Ratio method to external language model integration in automatic speech recognition \cite{mcdermott_densityratio} and hence refer to the model by this name.
This results in the following model combination:
\begin{align*}
    p\left(w_n \mid w_1^{n-1}, u_1^T, K^\prime \right) &\propto
    p_{\text{dm}}\left(w_n \mid w_1^{n-1}, u_1^T, K^\prime\right) \\
    &\cdot p_{\text{lm}}\left(w_n \mid w_1^{n-1}, u_1^T\right)^{\lambda_1} \\
    &\cdot p_{\text{ilm}}\left(w_n \mid w_1^{n-1}, u_1^T\right)^{-\lambda_2}
\end{align*}
The advantage of these two approaches is that the ungrounded response generation model can be trained on a large amount of dialog data without document grounding which allows a zero-shot domain transfer.
At the same time, however, this model might skew the distribution towards degenerate words that introduce inconsistencies w.r.t. the grounding if the scaling factor is too large and not influence the style sufficiently if it is too low.

\paragraph{Noisy Channel formulation}
Since it is not clear how shallow fusion influences the faithfulness of the generated responses, we seek to find a model that explicitly enforces faithfulness to document grounding.
Hence, we use Bayes Theorem to derive a noisy channel formulation for document-grounded response generation as follows:
\begin{align*}
    & \arg\max_{w} p(w \mid u_1^T, K') \\
    &= \arg\max_{w} p(w, u_1^T, K') \\
    &= \arg\max_{w} \underbrace{p(K' \mid w, u_1^T)}_{\text{channel model}} \cdot \underbrace{p(w \mid u_1^T)}_{\Substack{\text{response generation}\\ \text{model}}} 
\end{align*}
First of all, we can see that the advantage of having an ungrounded response generation model which can be trained on large amounts of textual data in the new domain without requiring document grounding is retained.
Furthermore, the channel model now encourages that the response explains the document grounding sufficiently well which could prevent the model from leaving out important details and mitigate the explaining-away effect \cite{liuPretrainingNoisyChannel2021a}.

However, decoding the noisy channel model directly is computationally intractable.
Hence, we use two different approximate decoding methods.
First of all, we experiment with \emph{reranking} generations obtained from a proposal model, for which we use the direct model.
That is, we first decode $k$ sequences from the direct model and then obtain the final response as the highest scoring sequence under the log-linear model combination
\begin{align}
    \label{reranking}
    \begin{split}
        \hat{w} = \arg\max_{w} & \log p\left(w \mid u_1^T, K^\prime\right)+ \\
        & \lambda_2 \cdot \log p\left(K^\prime \mid w, u_1^T\right)+ \\
        & \lambda_1 \cdot \log p\left( w \mid u_1^T\right)
    \end{split}
\end{align}
We interpolate with the direct model to encourage sequences with high direct model likelihood which has proven beneficial in other tasks \cite{yuNeuralNoisyChannel2017, liuPretrainingNoisyChannel2021a}.
While comparatively efficient, the method is limited by the proposal model, since the noisy channel formulation can only re-rank an n-best list of complete sequences.
For example, if all sequences contain false information, this error can not be corrected by the noisy channel model.

Therefore, we also derive a simple \emph{online decoding} algorithm that allows an early interaction between language and channel model.
At each timestep, the algorithm first decodes $k$ beams obtained from the direct model for each hypothesis on the beam.
Then, the beams are rescored using the log-linear model combination in Equation \ref{reranking}.
Since online decoding is based on partial sequences, we also need to ensure that the channel model is trained on partial inputs to not create a mismatch between train and test time.
Hence, as in \citet{liuPretrainingNoisyChannel2021a}, we truncate \(w\) according to a uniform distribution over all lengths from $1$ up to sequence length during training.

\section{Experiments}

The experiments have been done using HuggingFace Transformers \cite{wolfTransformersStateoftheArtNatural2020}, HuggingFace Datasets \cite{lhoest-etal-2021-datasets}, and Sisyphus \cite{peterSisyphusWorkflowManager2018}\footnote{Code is available at \url{https://github.com/dthulke/dstc10-track2}}.
All models were trained on Nvidia GTX 1080 Ti or RTX 2080 Ti GPUs.
For the evaluation, we use the same metrics as proposed by \citet{kimDomainAPIsTaskoriented2021} and that are used for the final ranking.
In the selection and generation subtasks which depend on the results of previous tasks, we evaluate the methods on the ground truth labels to facilitate comparability.

\subsection{Text Preprocessing}

\begin{table}
    \caption{Effect of different text preprocessing techniques in the detection task on the DSTC10 validation data.}
    \label{table:text_norm}
\begin{center}
\begin{tabular}{l|r} 
    \hline
    method & F1 \\\hline 
    baseline (RoBERTa-large) & 75.3 \\ 
    + lowercasing   & 78.4 \\
    \hphantom{+ }+ no punct.    & 79.7 \\ 
    \hphantom{+ }\hphantom{+ }+ numbers written out    & 83.7 \\
    \hphantom{+ }\hphantom{+ }\hphantom{+ }+ no abbrev.    & 84.1 \\ 
\end{tabular}
\end{center}
\end{table}

We experimented with the different proposed text processing strategies on the detection task.
\Cref{table:text_norm} shows the results on the DSTC10 test data.
We observed that each method gives a slight improvement in final performance.
Therefore, we decided to apply these pre-processing methods in the detection and selection tasks.

\subsection{Detection}

\begin{table}
    \caption{F1 scores of the detection subtask on the DSTC10 validation and test data.}
    \label{table:detection_results}
    \centering
    \begin{tabular}{l|r|r}
        \hline
        model & val & test  \\ \hline
        baseline (+ text preprocessing) & 84.8 & 84.1 \\
        + data augmentation & 91.9 & 85.1 \\
        \hphantom{+ }+ in-domain pretraining & 93.5 & 86.0 \\
        \hphantom{+ }\hphantom{+ }+ ASR n-best (weighted) & 94.5 & 86.5 \\
        \hphantom{+ }\hphantom{+ }+ ASR n-best (max) & 94.7 & 87.7 \\
        \hphantom{+ }\hphantom{+ }\hphantom{+ }+ DSTC9 test + DSTC10 val & - & 90.5 \\
        \hphantom{+ }\hphantom{+ }\hphantom{+ }\hphantom{+ }+ ensemble & - & 91.1 \\
    \end{tabular}
\end{table}

\Cref{table:detection_results} shows the results of our proposed methods for the detection task on the DSTC10 validation and test data.
First, augmenting the training data with additional samples generated from the knowledge base gave us a strong improvement on the validation and a small improvement on the test data.
The additional, in-domain pretraining of the RoBERTa model further improved the results by 1\%.
Next, we experimented with the two proposed ASR n-best strategies and observed better results with the max strategy.
Finally, we included the DSTC9 test and DSTC10 validation data into the training of the model and created an ensemble of different training runs.

\subsection{Selection}

\begin{table}
    \caption{R@1 scores of the selection subtask on the DSTC10 validation and test data.}
    \label{table:selection_results}
    \centering
    \begin{tabular}{l|r|r}
        \hline
        model & val & test  \\ \hline
        baseline (+ text preprocessing) & 71.2 & 70.0 \\
        + Beam Search & 74.0 & 73.5 \\
        \hphantom{+ }+ Taskmaster \& DSTC10 data & 78.8 & 76.3 \\
        \hphantom{+ }\hphantom{+ }+ in-domain pretraining & 83.7 & 77.0 \\
        \hphantom{+ }\hphantom{+ }\hphantom{+ }+ ASR n-best (max) & 79.8 & 77.7 \\
        \hphantom{+ }\hphantom{+ }\hphantom{+ }+ ASR n-best (weighted) & 81.7 & 77.7 \\
        \hphantom{+ }\hphantom{+ }\hphantom{+ }\hphantom{+ }+ DSTC9 test + DSTC10 val & - & 77.3 \\
        \hphantom{+ }\hphantom{+ }\hphantom{+ }\hphantom{+ }\hphantom{+ }+ ensemble & - & 77.6  \\
    \end{tabular}
\end{table}

\Cref{table:selection_results} shows the results of applying our proposed methods for the selection task on the DSTC10 validation and test data.
Using our proposed Beam Search approach instead of always taking the entity with the highest score results in an improvement of around 3\% absolute.
Further, training the domain and entity selection model on additional data from Taskmaster and DSTC10 Task 1 gives an additional improvement.
On the validation data, we observed slight degradations with our strategies to handle the ASR n-best list.
On the test set, this resulted in improvements.
We assume that the observed degradations can be attributed to the small size of the validation set.
In contrast to the detection task, including the DSTC9 test and DSTC10 validation data resulted in small degradations.
Finally, an ensemble of different training runs slightly improved the results again.

\subsection{Generation}

\begin{table}[H]
    \caption{Zero-shot transfer on the DSTC10 validation set.}
    \label{zero_shot}
    \centering
    \begin{tabular}{l|rrr}
        \hline
        Method            & \footnotesize{BLEU-1} & \footnotesize{METEOR}   & \footnotesize{ROUGE-L} \\ \hline \hline
        Direct model      & 21.1   & 24.4     & 19.6    \\
        + Shallow Fusion    & 20.6   & 25.5     & \textbf{22.9}    \\
        + Density Ratio& 21.2    & \textbf{26.0}     & \textbf{22.9} \\
        + Noisy Channel$_\text{Onl.}$& \textbf{22.5}     & 24.6     & 22.3
    \end{tabular}
\end{table}

Similar to previous work \cite{thulkeEfficientRetrievalAugmented2021}, we fine-tuned BART large for all sequence-to-sequence models discussed in this section.
Furthermore, the ungrounded response generation model is always trained on the spoken two-person subset of Taskmaster-1 \cite{byrneTaskmaster1RealisticDiverse2019}, Taskmaster-2, CCPE \cite{radlinski-etal-2019-ccpe}, as well as the indicated DSTC splits.
We study different settings for the generation task.
In the first setting, we train the models only on the written dialog training set from the DSTC9 challenge.
Then, we perform a zero-shot evaluation concerning the spoken domain dialogs of the DSTC10 validation set.
Table \ref{zero_shot} shows results obtained with different models in this setting.
We can see that both the shallow fusion and density ratio approach give slight improvements.
By using noisy channel online decoding with the direct model as proposal model we obtain similar results.
However, these methods require substantial additional computational effort and the increased number of parameters from combining multiple models might also contribute to the improvements.

In general, we note that the generations in the zero-shot setting are substantially different from the references on the spoken test set and particularly contain follow-up questions such as "Do you need anything else?", which are not contained in the responses of the test data.
We have also experimented with removing these phrases in the training data which however brought degradations in terms of automatic metrics, mainly due to the included length penalty.
The test data contains different rather colloquial phrases such as the prefix "sure let me check that for you".

\begin{table}
    \caption{Trained on DSTC9 train and 10-fold cross-validation on DSTC10 val.}
    \label{table:few_shot}
    \centering
    \begin{tabular}{l|rrr}
        \hline
        Method            & \footnotesize{BLEU-1} & \footnotesize{METEOR}   & \footnotesize{ROUGE-L} \\ \hline \hline
        \footnotesize{Direct Model}      & 40.6   & 49.0     & 44.2    \\
        \footnotesize{+ Noisy Channel}   & 41.9   & 50.7     & 46.0\\
        \footnotesize{+ Style Token}     & 42.3   & 50.2     & 45.7         \\
        \footnotesize{\hphantom{+ }+ Noisy Channel$_\text{Onl.}$}   & \textbf{42.5}   & \textbf{51.3}     & \textbf{46.2} \\
    \end{tabular}
\end{table}

To also adapt the style of these phrases in the response, we experiment with training our model on the DSTC10 validation data.
We split the data in 10 cross validation splits to still be able to evaluate the resulting models on the full validation set.
The results are highlighted in Table \ref{table:few_shot}.
We can see that using a style token gives good improvements.
Furthermore, the noisy channel model with online decoding gives further improvements for both proposal models.

\begin{table}[H]
    \caption{Trained on DSTC9 train and DSTC10 val, evaluated on DSTC10 test.}
    \label{table:generation_test}
    \centering
    \begin{tabular}{l|rrr}
        \hline
        Method            & \footnotesize{BLEU-1} & \footnotesize{METEOR}   & \footnotesize{ROUGE-L} \\ \hline \hline
        Direct Model      & 43.1   & 49.8     & 48.0    \\
        + Shallow Fusion    & 43.4   & 50.1     & 48.2    \\
        + Style Token  & 43.8   & 50.3     & 48.7    \\
        \hphantom{+ }+ Noisy Channel$_\text{Re.}$  & 44.3   & 51.2     & 48.8    \\
        \hphantom{+ }+ Noisy Channel$_\text{Onl.}$& \textbf{45.7} & \textbf{52.0}     & \textbf{50.2}    \\
    \end{tabular}
\end{table}

Finally, Table \ref{table:generation_test} shows our results on the DSTC10 test data.
All models were trained on the DSTC9 training data as well as both validation sets of DSTC9 and 10, i.e. a small amount of in-domain training data containing ASR errors and spontaneous speech.
In this setting, there is only a minor improvement by interpolating with a spoken response generation model.
Explicit conditioning on the style using a style token gives stronger improvements without necessitating the training of an additional model and is thus our preferred method for obtaining proposals for the noisy channel model which shows improvements with both decoding strategies.
The online decoding method forms our best model and our submission to the shared task.
This indicates that the early interaction of response generation and channel model can indeed improve performance and at the same time that the reranking approach is too limited by the proposal model.
We discuss the improvements observed by the model in Section \ref{error_analysis}.
We, however, need to note that the model has more parameters than the direct model such that some of the improvements might be attributed to an increase in modeling capacity.
Nevertheless, as shown by \citet{liuPretrainingNoisyChannel2021a}, noisy channel formulations can still outperform the direct modeling approach with the same amount of parameters for task-oriented dialog.
Hence, we do not study this particular aspect in our paper.
\subsection{Challenge Results}

\begin{table*}
    \caption{Final results of the top 5 out of 16 teams ranked according to the human evaluation.}
    \label{table:main_results}
    \centering
    \begin{tabular}{l|r|r|r|r|r|r|r}
        \hline
         & detection  &                selection                     & \multicolumn{3}{l|}{generation}& \multicolumn{2}{l}{human evaluation}  \\
                                   & {F1}                                           & {R@1}                              & {BLEU-1}        & {METEOR}        & {ROUGE-L} & {Accuracy} & {Appropriateness} \\ \hline \hline
         baseline                   & 79.5                                      & 45.8                          & 11.5         & 12.2          & 11.4 & 2.74 & 2.79 \\ \hline
         Team 10                    & 92.3                                       & \textbf{79.3}         & 16.2         & 21.0         & 21.9 & \textbf{3.49} & \textbf{3.35} \\ \hline
         Team 4                     & 91.8                                       & 74.8         & 33.8         & 40.7         & 38.7 & 3.34 & 3.30  \\ \hline
         Team 8 (our)              & 91.1                                       & 71.0         & \textbf{40.1}         & \textbf{46.0}         & \textbf{44.1} & 3.34 & 3.26  \\ \hline
         Team 14                    & \textbf{92.4}                                       & 62.0         & 27.1         & 31.7         & 31.8 & 3.29 & 3.28  \\ \hline
         Team 2                    & 90.4                                       & 69.3         & 37.3         & 43.9         & 41.2 & 3.29 & 3.22  \\ \hline \hline
         \multicolumn{6}{l|}{ground truth}                 & 3.58 & 3.48  \\ \hline
    \end{tabular}
\end{table*}

\Cref{table:main_results} shows the official results of DSTC10 Track 2 Task 2 of the best five teams according to the human evaluation.
The baseline system is the original baseline system proposed by \citet{kimDomainAPIsTaskoriented2020} for DSTC9.
In contrast to previous experiments, models for the selection and generation subtasks are applied to the outputs of the previous tasks.
In total, 16 teams participated in the challenge.

\paragraph{Automatic Evaluation}
Our best system achieved 4th place in the detection (F1) and selection (R@1) subtasks and 1st place in the generation subtask.
In the official ranking according to the automatic metrics, our system achieved 1st place as the generation metrics have a higher influence on the final ranking than the other tasks.

\paragraph{Human Evaluation}
For the 8 best systems according to the automatic evaluation an additional human evaluation was conducted by the task organizers.
Crowd workers were asked to rate the accuracy and appropriateness of a system response on a scale from 1 to 5.
For the former, they had access to the reference knowledge document and should rate the accuracy of the reference response.
To rate the appropriateness, they only had access to the dialog context and should rate how well the response fits into the dialog.
As in the previous year's evaluation \cite{kimDomainAPIsTaskoriented2021}, the results in the selection task have the strongest correlation to human judgments.
Additionally, we observe that there is no apparent correlation between the automatic scores in the generation task and the human judgements.
For example, Team 10 ranks first in the human evaluation even though they have the lowest scores in the automatic generation metrics among the top five teams.
One possible explanation is that crowd workers did not prefer responses in spoken style, even though they should fit more naturally in the spoken conversations.

\subsection{Error Analysis}
\label{error_analysis}
\begin{table*}[ht]
    \centering
    \caption{Comparison of responses generated by the noisy channel and direct model.}
    \label{table:comparison_outputs}
    \begin{tabular}{p{0.15\linewidth}p{0.25\linewidth}p{0.48\linewidth}}
    \toprule[.1em]
    Last turn $u_T$ & selected document & Model: Generated response $u_{T+1}$\\
    \midrule[.1em]
    \multirow{2}{=}{o great and do you know if the taj happens to uh allow smoking} & \multirow{2}{=}{Is there a smoking area at Taj Campton Place?}& \emph{Baseline}: yeah let me check that for you all right so it does say that smoking is allowed here\\
    \addlinespace 
    & Yes, the Taj Campton Place has a designated smoking area. & \emph{Noisy Channel}: yeah let me check that for you all right so they do have a designated smoking area at this location\\
    \midrule[.1em]
    \multirow{2}{=}{two stars ok do you know if they have a reservation or parking} & \multirow{2}{=}{Do I need to make a reservation for parking in advance?}& \emph{Baseline}: yeah let me go ahead and see here all right so it's saying that they do not have a reservation or parking at this location\\
    \addlinespace 
    & Parking spots are available on a first come, first served basis. No reservation is available now. & \emph{Noisy Channel}: yeah let me go ahead and see here all right so it's saying that they do not have a reservation but parking spots are available on a first come, first served basis\\
    \midrule[.1em]
    \multirow{2}{=}{can i bring a dog there} & \multirow{2}{=}{Are dogs allowed?\\}& \emph{Baseline}: they do not\\
    \addlinespace 
    & Aquarium Of The Bay does not allow dogs. & \emph{Noisy Channel}: i'm sorry they do not allow dogs at this location\\
    \midrule[.1em]
    \multirow{2}{=}{oh great and what's this place lt ike if p m b m} & \multirow{2}{=}{What is the atmosphere like?\\}& \emph{Baseline}: yeah let me go ahead and search that all right so it's not saying that you can't bring a dog there but it does say that it's a classy and casual place\\
    \addlinespace 
    & It has a classy and casual ambience with a great outdoor view of the water. & \emph{Noisy Channel}: let me go ahead and search that all right so it's listed here as classy and casual with a great outdoor view of the water\\
    \midrule[.1em]
    \multirow{2}{=}{l that sound great ill just wondering do they accept accept the scoffer card} & \multirow{2}{=}{What type of payment method do you accept?\\}& \emph{Baseline}: yeah let me go ahead and check if that's listed anywhere ok so they do accept all major credit cards including scoffer card\\
    \addlinespace 
    & The Nob Hill Motor Inn accepts, Visa, Mastercard, American Express and Discover. & \emph{Noisy Channel}: yeah let me go ahead and check that for you. ok so they do accept all major credit cards including the scoffer card\\
    \end{tabular}
\end{table*}

Error analysis of the generation models shows different patterns.
First of all, we observe that all models sometimes do not answer the previous user question but rather one in the dialog context.
We hypothesize that the model can not make sense of the previous turn, also in combination with the knowledge document, and then generates a false response.
This would suggest modeling only the previous user turn, which however did not result in improvements in terms of automatic metrics in our initial experiments.
Furthermore, the model often repeats ASR errors, especially when an entity is not recognized, and generates an answer containing the false transcription such as in the last row of Table \ref{table:comparison_outputs} as an example, where both models repeat the word "scoffer".
When comparing the direct model with the noisy channel formulation, we can see that the direct model sometimes leaves out important details which the noisy channel model retains.
For example, in the first row of Table \ref{table:comparison_outputs}, we can see that the former only mentions that smoking is allowed while the latter model points out designated areas.
Furthermore, the direct model sometimes fails to interpret the document correctly, for example in line three where the direct model incorrectly says that no parking is available which the noisy channel method is able to identify correctly.
It is also worth noting that such effects depend on the scaling factors of the noisy channel model.
In our experiments, too much influence of the response generation model led to hallucinations and too much influence of the channel model led to the model mostly repeating the document.

\section{Related Work}
It has been shown that models trained on written dialogs generalize poorly towards spoken dialogs with ASR errors \cite{Gopalakrishnan2020AreNO, kimHowRobustEvaluating2021}.
Thus, different approaches of increasing the robustness of Spoken Language Understanding models to ASR errors have been proposed.
Traditionally, many systems have used confusion networks \cite{tur02_icslp, hendersonDiscriminativeSLU2012} to consider more than one ASR hypothesis.
A different method is to already include automatic speech recognition errors in the training data.
For example, \citet{Schatzmann2007error} introduce ASR errors based on an n-gram confusion matrix obtained after Levenshtein alignment.
\citet{FazelZarandi2019InvestigationOE} propose to use vocoders to create spoken sequences from textual training samples, adding noise and then transcribing the samples with an ASR model to obtain perturbed training data.
Simpler transformations such as removing punctuation \cite{Gopalakrishnan2020AreNO} have also been proposed.
The inverse transformation of correcting ASR errors has also been explored, for example in \citet{Weng2020JointCM}.

Document-grounded dialog datasets have so far focused on the written domain.
For example, Wizard-of-Wikipedia \cite{dinanWizardWikipediaKnowledgePowered2018} consists of open-domain dialogs grounded by Wikipedia articles, and the doc2dial dataset \cite{fengDoc2dialGoalOrientedDocumentGrounded2020} consists of task-oriented dialogs grounded by long documents.
In this paper, we are concerned with a continuation of the task "Beyond Domain APIs: Task-oriented Conversational Modeling with Unstructured Knowledge Access" hosted at DSTC9 \cite{kimDomainAPIsTaskoriented2021} which contains dialogs grounded in FAQ documents.
Thus, we review different methods introduced for the tasks of Turn Detection, Knowledge Selection, and Response Generation.
\citet{heLearningSelectExternal2021} propose to model the first task by deciding whether a knowledge document or schema description obtained from MultiWOZ is more likely to be sought by the user.
In the first case, the most likely knowledge document is selected and in the latter case, the turn is deemed not knowledge-seeking.
Furthermore, similar to \citet{jinCanBeFurther2021} the authors propose different strategies to sample negative training examples, such as sampling documents from the same domain or entity.
\citet{tangRADGERelevanceLearning2021} propose to select negatives by first training a model on the selection task with negatives sampled from the same entity or domain and then taking the documents likely to be confused under the model as negatives to train a stronger model.
While this forms an explicit negative sampling, \citet{thulkeEfficientRetrievalAugmented2021} explore to fine-tune the selection models end-to-end with the response generation task by using a retrieval augmented model \cite{guuREALMRetrievalAugmentedLanguage2020, lewisRetrievalAugmentedGenerationKnowledgeIntensive2020}, where the marginalization can be seen as an implicit batching of hard negatives.
Furthermore, the authors propose to use a hierarchical selection approach and formulate knowledge selection as a metric learning problem using bi-encoders, similar to \citet{karpukhinDensePassageRetrieval2020}.
Finally, \citet{miGeneralizedModelsDomain2021} and \citet{kimHowRobustEvaluating2021} propose different data augmentation methods to augment the training data by unseen knowledge documents.

In general, pretrained language models have found great success to perform the aforementioned tasks.
Notably, RoBERTa \cite{liuRoBERTaRobustlyOptimized2019} has often been used for the sequence classification tasks, while BART \cite{lewisBARTDenoisingSequencetoSequence2020} has often been used for response generation, as is suggested by their corresponding pretraining objectives.

Parallel to our work on the Density Ratio model, \citet{liu-etal-2021-dexperts} applied a similar method to controlled generation.

The Noisy Channel decomposition \cite{shannonnoisychannel} has a long history in different language technology tasks, such as machine translation \cite{brown-etal-1993-mathematics} or automatic speech recognition \cite{bahl_jelinek}.
With the advent of deep learning, modeling these tasks discriminatively has often been the preferred choice.
Nevertheless, recently neural noisy channel modeling has been explored for different tasks, such as machine translation \cite{yuNeuralNoisyChannel2017, yeeSimpleEffectiveNoisy2019, jeanLogLinearReformulationNoisy2020, subramanian2021nvidia}, few-shot text classification \cite{minNoisyChannelLanguage2021}, and dialog \cite{liuPretrainingNoisyChannel2021a}.

\section{Conclusion}
We have proposed different methods of adapting document-grounded dialog models to noisy transcriptions of spoken conversations.
Notably, we achieve significant improvements in the tasks of turn detection and knowledge selection with rather simple text preprocessing steps.
For both tasks, we also propose to use multiple ASR hypotheses which can prevent some errors, for example when an important entity is present only in a few hypotheses.
Furthermore, we highlight a data augmentation method to create samples for new knowledge documents by adapting dialogs from MultiWOZ.
For the knowledge selection task, we propose a beam search for hierarchical selection models and pretrain the model on Taskmaster.
We also explore different methods to improve the domain adaption of response generation models.
In particular, we show the benefits of using a noisy channel factorization to leverage ungrounded dialog data and enforce the consistency of responses regarding their document grounding.

In the future, it would be interesting to explore a tighter integration with speech recognition models.
This would allow for better error propagation and to make use of paralinguistic features like emphasis or emotion.
Furthermore, we plan to study alternative online decoding algorithms for the noisy channel model to alleviate the strong dependency on the proposal model and especially create more diverse proposals.
We also want to gain a better understanding of the model and its differences to the direct modeling approach.
With this in mind, evaluation strategies that capture aspects such as diversity and especially faithfulness are of interest.

\section*{Acknowledgments}
This work has received funding from the European Research Council (ERC) under the European Union's Horizon 2020 research and innovation programme (grant agreement No 694537, project ``SEQCLAS''). The work reflects only the authors' views and the European Research Council Executive Agency (ERCEA) is not responsible for any use that may be made of the information it contains.

\bibliography{david, nico}

\end{document}